\documentclass[11pt,twocolumn]{article}

\usepackage[margin=0.8in]{geometry}
\usepackage{times}
\usepackage{epsfig}
\usepackage{graphicx}
\usepackage{float}
\usepackage{amsmath}
\usepackage{amssymb}
\usepackage{amsfonts}
\usepackage{siunitx}
\usepackage{algorithm2e}
\usepackage{caption}
\usepackage{dsfont}
\usepackage{booktabs}
\usepackage{subfigure}
\usepackage{tikz}
\usepackage{cite}
\usepackage{color}
\usepackage{textcomp}
\usepackage{caption}
\usepackage[hyphens]{url}
\usepackage[pagebackref=true,breaklinks=true,letterpaper=true,colorlinks,bookmarks=false]{hyperref}

\makeatletter
\renewcommand{\@biblabel}[1]{\quad#1.}
\makeatother

\date{}

\begin{document}

\begin{flushleft}
{\Large
\textbf{Machine learning of hierarchical clustering to segment 2D and 3D images}
}

Juan Nunez-Iglesias$^{1,\dagger}$,
Ryan Kennedy$^{2}$,
Toufiq Parag$^{1}$,
Jianbo Shi$^{2}$,
Dmitri B. Chklovskii$^{1}$
\\
\bf{1} Janelia Farm Research Campus, Howard Hughes Medical Institute, Ashburn, VA, USA
\\
\bf{2} Department of Computer and Information Science, University of Pennsylvania, Philadelphia, PA, USA
\\
\bf{$\dagger$} Email: jni@janelia.hhmi.org
\end{flushleft}

\section*{Abstract}
We aim to improve segmentation through the use of machine learning tools during region agglomeration. 
We propose an active learning approach for performing hierarchical agglomerative segmentation from superpixels. 
Our method combines multiple features at all scales of the agglomerative process, works for data with an arbitrary number of dimensions, and scales to very large datasets.
We advocate the use of variation of information to measure segmentation accuracy, particularly in 3D electron microscopy (EM) images of neural tissue, and using this metric demonstrate an improvement over competing algorithms in EM and natural images.

%%% INTRODUCTION %%%
%\input{intro.tex}

\section{Introduction}

Image segmentation, a fundamental problem in computer vision, concerns the division of an image into meaningful constituent regions, or segments.

\begin{figure}[!h]
\centering
\includegraphics[width=3.27in]{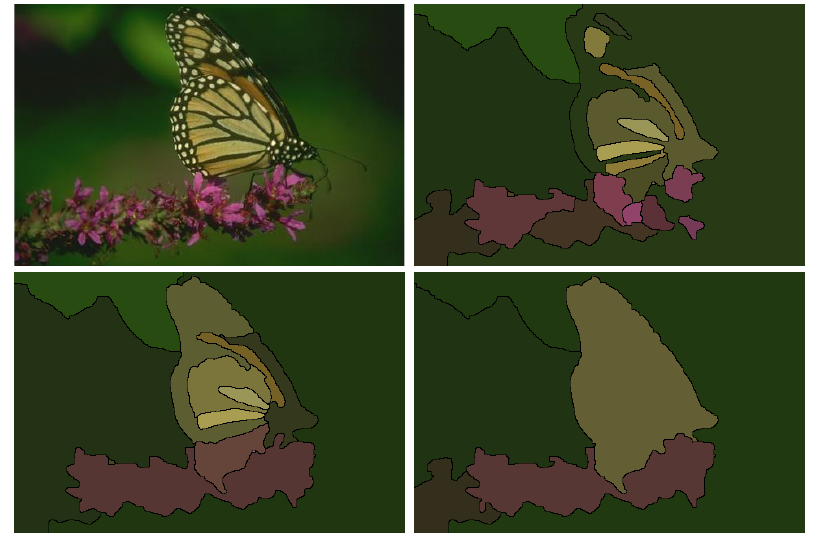}
\caption{{\bf Illustration of the advantages of our approach.} {\it Top left}: Input image. {\it Top right}: segmentation using only a boundary map \cite{arbelaez10}. {\it Bottom left}: using multiple cues with a single level of learning. {\it Bottom right}: using multiple cues with our agglomerative learning method.  }
\label{fig:butterfly-example}
\end{figure}

In addition to having applications in computer vision and object recognition (Figure \ref{fig:butterfly-example}), it is becoming increasingly essential for the analysis of biological image data.
Our primary motivation is to understand the function of neuronal circuits by elucidating neuronal connectivity \cite{Anderson:2009fz,Chklovskii:2010df}.
In order to distinguish synapses and follow small neuronal processes, resolutions of \texttildelow 10nm are necessary in 3D and provided only by electron microscopy (EM).
On the other hand, individual neurons often extend over millimeter ranges.
This disparity of scales results in huge image volumes and makes automated segmentation an essential part of neuronal circuit reconstruction.

\begin{figure*}[!ht]
\centering
\includegraphics[width=6.83in]{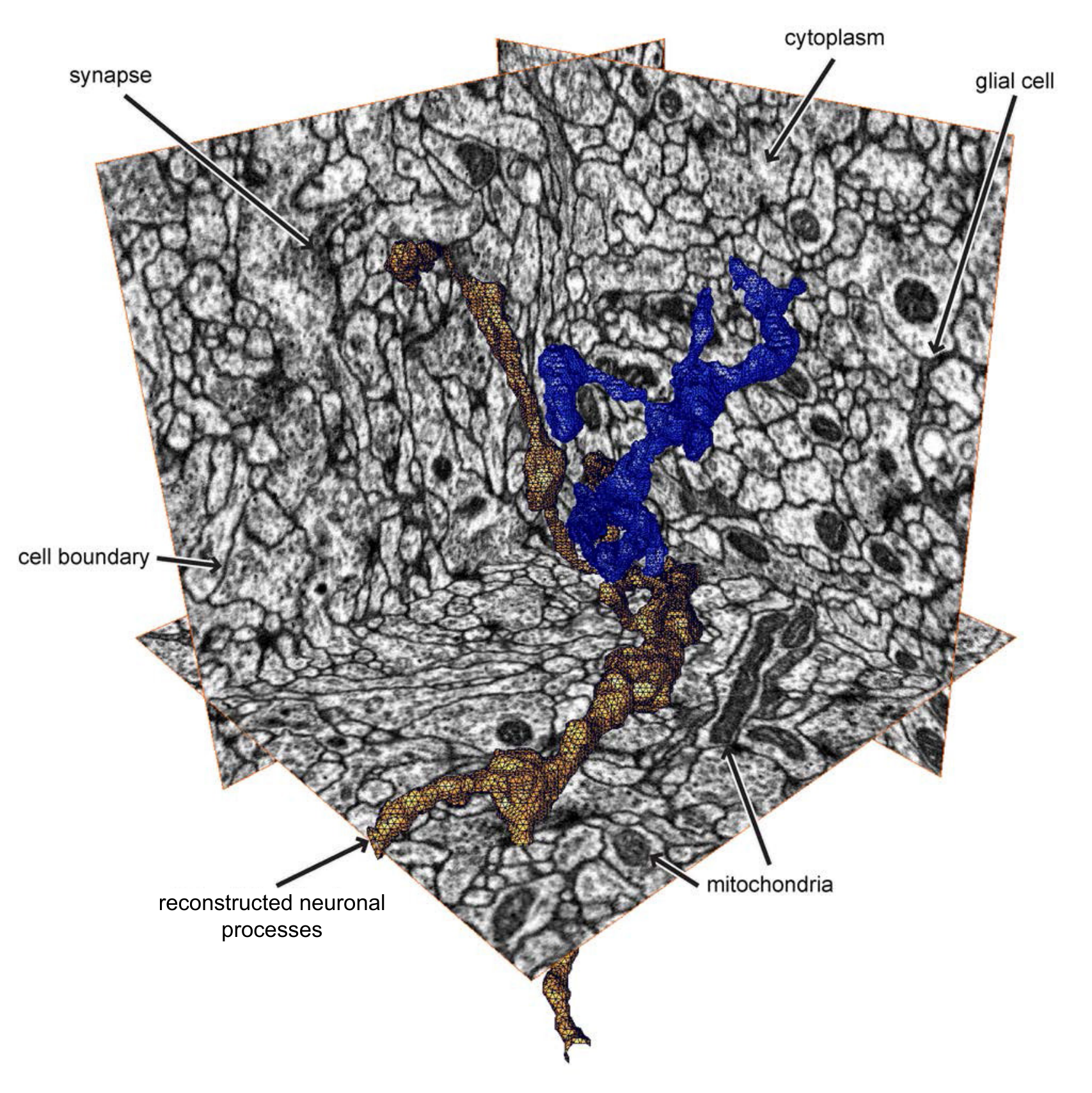}
\caption{{\bf Representative 3D EM data and sample reconstructions.} Note that the data is isotropic, meaning it has the same resolution along every axis. The goal of segmentation here is to partition the volume into individual neurons, two of which are shown in orange and blue. The volume is densely packed by these thin neuronal processes taking long, tortuous paths. }
\label{fig:em-images}
\end{figure*}

Additionally, automated segmentation of EM images presents significant challenges compared to that of natural images (Figure \ref{fig:em-images}), including identical textures within adjacent neurons, mitochondria and vesicles within cells that look (to a classifier) similar to the boundaries between cells, and elongated, intertwined shapes where small errors in boundary detection result in large errors in neuron network topology.
The methods we introduce here, however, are generally applicable and extend to images of arbitrary dimension, which we demonstrate by segmenting both EM data and natural image data.

A common approach in the field is to perform oversegmentation into small segments called \emph{superpixels}, and then to merge these into larger regions \cite{Ren:2003jg, arbelaez10}.
A merging algorithm consists of a merging criterion, or policy, that determines which merges are most likely, and a merging strategy, that determines how to merge segments (for example, through simulated annealing \cite{Ren:2003jg}, probabilistic graphical models \cite{Andres:2011is}, or hierarchical clustering \cite{jain2011lash}).
Often, much effort is devoted to the generation of a pixel-level boundary probability map by training a classifier that predicts boundaries between objects from pixel-level features \cite{martin_learning_2004, dollar2006supervised, Turaga:2009, arbelaez10, Jain:2010, Jurrus:2010fw}. 
Meanwhile, oversegmentation and agglomeration are performed in a straightforward fashion, for example using watershed \cite{Vincent:1991} to generate superpixels, and the mean boundary probability over the contour separating adjacent superpixels \cite{arbelaez10} as the merge criterion.
Boundary mean has been a relatively effective merge priority function for hierarchical agglomeration because every merge results in longer boundaries along adjacent regions.
Therefore, as the agglomeration proceeds, the mean becomes an increasingly reliable estimate of the merge probability.

We hypothesized that agglomeration could be improved by using more information than just the boundary mean, despite the latter's desirable characteristics.
A priority function could draw from many additional features, such as boundary variance and region texture.
Using training data in which pairs of superpixels have been labeled as ``merge'' or ``don't merge'', we could then apply machine learning techniques to predict from those features whether two superpixels should be merged.
With that simple approach, however, we found that the guaranteed effectiveness of the mean could easily disappear.
Similarly to the case with the boundary mean, the region sizes progressively increase and so does the amount of evidence for or against a merge.
However, we could encounter a combination of features for which we had no training data.

To get around this problem, we developed an active learning paradigm that generates training examples across every level of the agglomeration hierarchy and thus across very different segment scales.
In active learning, the algorithm determines what example it wants to learn from next, based on the previous training data.
For agglomerative segmentation, we ask the classifier which two regions it believes should be merged, and compare those against the ground truth to obtain the next training example.
By doing this at all levels of the agglomeration hierarchy, we ensure that we have samples from all parts of the feature space that the classifier is likely to encounter.

Past learning methods either used a manual combination of a small number of features \cite{arbelaez10, grundmann2010efficient}, or they used more complex feature sets but operated only on the scale of the original superpixels \cite{andres08, chengmulti}.
(We discuss two notable exceptions \cite{Funke:2012uv,Liu:2012ba} in Section \ref{sec:discussion}.)
We instead learn by performing a hierarchical agglomeration while comparing to a gold standard segmentation.
This allows us to obtain samples from region pairs at all scales of the segmentation, corresponding to levels in the hierarchy.
Although Jain \textit{et al.} independently presented a similar approach called LASH \cite{jain2011lash}, there are some differences in our approach that yield some further improvements in segmentation quality, as we explain later.

We describe below our method for collecting training data for agglomerative segmentation.
Throughout a training agglomeration, we consult a human-generated gold standard segmentation to determine whether each merge is correct.
This allows us to learn a merge function at the many scales of agglomeration.
We show that our learned agglomeration outperforms state of the art agglomeration algorithms in natural image segmentation (Figure \ref{fig:butterfly-example}).

To evaluate segmentations, we advocate the use of variation of information (VI) as a metric and show that it can be used to improve the interpretability of segmentation results and aid in their analysis.

The ideas in this work are implemented in an open-source Python library called \href{https://github.com/janelia-flyem/gala}{Gala} that performs agglomeration learning and segmentation in arbitrary dimensions.

\begin{figure*}[!ht]
\centering
\includegraphics[width=6.83in]{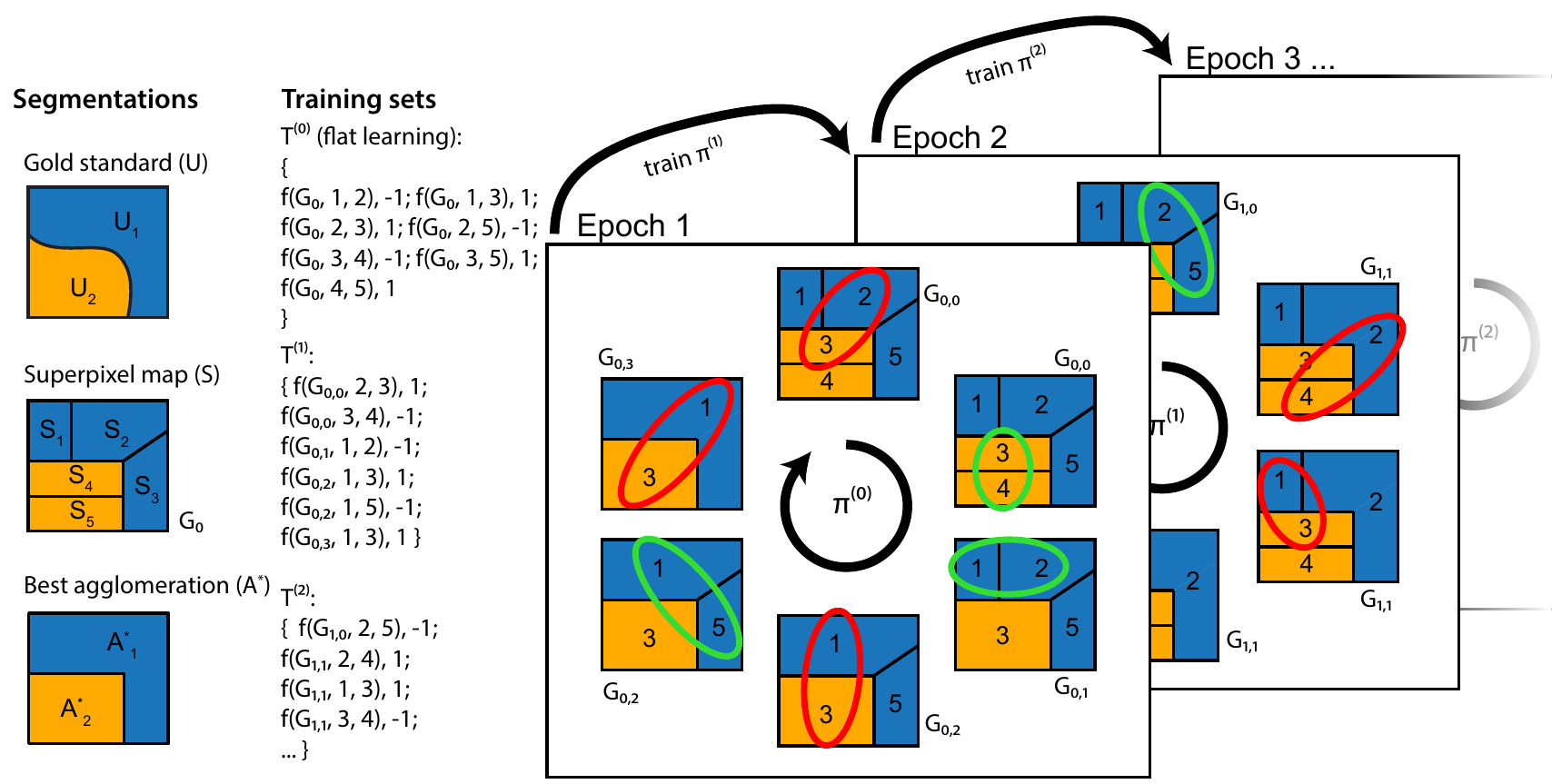}
\caption{{\bf Schematic of our approach.} {\it First column: }A 2D image has a given gold standard segmentation $U$, a superpixel map $S$ (which induces an initial region adjacency graph, $G_0$), and a ``best'' agglomeration given that superpixel map $A^*$. {\it Second column: } Our procedure gives training sets at all scales. ``f'' denotes a feature map. $G_{i, j}$ denotes graph agglomerated by policy $\pi^{(i)}$ after $j$ merges. Note that $j$ only increases when we encounter an edge labeled $-1$. {\it Third column:} We learn by simultaneously agglomerating and comparing against the best agglomeration, terminating when our agglomeration matches it. The highlighted region pair is the one that the policy, $\pi^{(k)}$, determines should be merged next, and the color indicates the label obtained by comparing to $A^*$. After each training epoch, we train a new policy and undergo the same learning procedure. For clarity, in the second and third columns, we abbreviate $A_i$ with just the index $i$ in the second and third arguments to the feature map. For example, $f(G_{0,0}, 2, 3)$ indicates the feature map from graph $G_{0,0}$ and edge $(v_2, v_3)$, corresponding to regions $A_2$ and $A_3$. }
\label{fig:agglo-learning}
\end{figure*}

%%% METHODS %%%
% \input{methods.tex}

\section{Methods}

\subsection{Active learning of agglomeration}
\label{methods:agglo-learning}
        The method described below is illustrated and summarized in Figure \ref{fig:agglo-learning}.

        Let $I \in \mathbb{R}^{n}$ be an input image of dimension $d$ having $n$ pixels.
        (Throughout the text, we will use ``pixel'' and ``voxel'' interchangeably.)
        We assume an initial oversegmentation $S$ of $I$ into $m << n$ ``superpixels'', $S = \{S_1,\dots,S_m\}$, defined as disjoint sets of connected pixels that do not substantially cross true segment boundaries.
        An agglomerative segmentation of the image is defined by a grouping $A = \{A_1, \dots, A_p\}$ of disjoint sets of superpixels from $S$.
        It is a testament to the power of abstraction of agglomerative methods that we will no longer use $d$, or $n$ in what follows.

        There are many methods to obtain $A$ from $I$ and $S$.
        We chose the framework of hierarchical agglomeration for its inherent scalability: each merge decision is based only on two regions.
        For this method we require two definitions: a region adjacency graph (RAG) and a merge priority function (MPF) or policy.

        The RAG is defined as follows.
        Each node $v_i$ corresponds to a grouping $A_i$ of superpixels, where we initialize $A_i \equiv \{S_i\}$, for $i = 1, \dots, m$.
        An edge $e_{i,j}$ is placed between $v_i$ and $v_j$ if and only if a pixel in $A_i$ is adjacent to a pixel in $A_j$.

        We then define the merge priority function (MPF) or policy $\pi : \{\mathcal{G}, V \times V\} \mapsto \mathcal{D} \subseteq \mathbb{R}$, where $\mathcal{G}$ is the set of RAGs and $V$ is the set of nodes beloging to a RAG.
        $\mathcal{D}$, the range of the policy, is typically $[0, 1]$, but could be any totally ordered set.
        Hierarchical agglomeration is the process of progressively merging nodes in the graph in the order specified by $\pi$.
        When two nodes are merged, the set of edges incident on the new node is the union of their incident edges, and the MPF value for those edges is recomputed.
        (A general policy might need to be recomputed for \emph{all} edges after a merge, but here we consider only local policies: the MPF is only recomputed for edges for which one of the incident nodes has changed.)

        The mean probability of boundary along the edge is but one example of a merge priority function.
        In this work, we propose finding an optimal $\pi$ using a machine learning paradigm.
        To do this, we decompose $\pi$ into a feature map $f : \{\mathcal{G}, V \times V\} \mapsto \mathbb{R}^q$ and a classifier $c : \mathbb{R}^q \mapsto [0, 1]$.
        Then take $\pi = c \circ f$, and the problem of learning $\pi$ reduces to three steps: finding a good training set, finding a good feature set, and training a classifier.
        In this work, we focus on the first question.
        The method we describe in the following paragraphs is summarized in Figure \ref{fig:agglo-learning}.

        We first define the optimal agglomeration $A^*$ given the superpixels $S$ and a gold standard segmentation $U$ by assigning each superpixel to the ground truth segment with which it shares the most overlap:

        \begin{eqnarray}
            A^*(S, U) & = & \left\{ A_i^* \right\}_{i=1}^{|U|} \\
            \textrm{ where }
            A_i^* & = & \left\{S_j : 
                    i = \arg\max_{k=1,\dots,|U|} 
                        {|S_j \cap U_k|}
                    \right\}_{j=1}^{|S|} \textrm{ .}
        \end{eqnarray}

        From this, we can work out a label between two regions: $-1$ or ``should merge'' if both regions are subsets of the same gold standard region, $1$ or ``don't merge'' if each region is a subset of a different gold standard region, and $0$ or ``don't know'' if either region is not a subset of any gold standard region:

        \begin{eqnarray}
        \ell(A^{*}, A_i, A_j) = 
            \begin{cases}
                -1, \textrm{ if } 
                    A_i \subseteq A^{*}_u, A_j \subseteq A^{*}_u
                    \textrm{ for some $u$ } \\
                1, \textrm{ if } 
                    A_i \subseteq A^{*}_u, A_j \subseteq A^{*}_v
                    \textrm{ for some $u \neq v$ } \\
                0, \textrm{ otherwise}
            \end{cases}
        \end{eqnarray}

        Now, given an initial policy $\pi^{(0)}$ and a feature map $f$, we can obtain an initial agglomeration training set as follows:
        Start with an initially empty training set $T$.
        For every edge $(u, v)$ suggested by $\pi^{(0)}$, compute its label $\ell_{u,v}$.
        If it is $-1$, add the training example $\{f(G, u, v), \ell_{u, v}\}$ to $T$ and merge nodes $u$ and $v$.
        Otherwise, add the training example but do not merge the two nodes.
        Repeat this until the agglomeration induced by the RAG $G$ matches $A^{*}$, and use $T$ to train a classifier $c$.
        We call this loop a training epoch.

        After epoch $k = 1, \dots, K$, we obtain a classifier $c^{(k)}$ that induces a policy $\pi^{(k)} = c^{(k)} \circ f$.

        There remains the issue of choosing a suitable initial policy.
        We found that the mean boundary probability or even random numbers work well, but, to obtain the fastest convergence, we generate the training set consisting of every labeled edge in the initial graph (with no agglomeration), $T^{(0)} = \left\{ ( f(G, e), \ell_{e} ) \right\}_{e \in E}$, and an initial policy is given by the classifier trained on this ``flat learning'' set.
        
\subsection{Cues and features}
\label{sec:features}
    In this section, we describe the feature maps used in our work.
    We call primitive features ``cues'', from which we compute the actual features used in the learning.
    We did not focus on these maps extensively, and expect that these are not the last word with respect to useful features for agglomerative segmentation learning.

    For natural images, we use the gPb oriented boundary map \cite{arbelaez10} and a texton map \cite{Brendel:2010wa}.
    For any feature calculated from gPb, the probability associated with an edge pixel was taken from the oriented boundary map corresponding to the orientation of the edge pixel.
    We calculated each edge pixel's orientation by fitting line segments to the boundary map and calculating the orientation of each line segment. By fitting line segments we are able to accurately calculate the orientation of each edge pixel, even near junctions where the gradient orientation is ambiguous \cite{arbelaez10}. 
    In addition, we use a texton cue that includes L*a*b* color channels as well as filter responses to the MR8 filter bank \cite{varma2005statistical, brendel2010segmentation}.
    The textons were discretized into 100 bins using the k-means algorithm.  

    For EM data, we use four separate cues: a probability map of cell boundaries, cytoplasm, mitochondria, and glia. 
    Mitochondria were labeled by hand using the active contours function in the ITK-SNAP software package \cite{py06nimg}.
    Boundaries and glia were labeled using the manually proofread segmentation in Raveler \cite{Chklovskii:2010df}, with cytoplasm being defined as anything not falling into the prior three categories.
    Our initial $500 \times 500 \times 500$ voxel volume was divided into 8 $250 \times 250 \times 250$ voxel subvolumes.
    To obtain the pixel-level probability map for each subvolume, we trained using the fully labeled 7 other subvolumes using Ilastik \cite{ilastik_0_5} and applied the obtained classifier.
    Rather than using all the labels, we used all the boundary labels (\texttildelow 10M total) and smaller random samples of the cytoplasm, mitochondria, and glia labels (\texttildelow 1M each).
    We found that this resulted in stronger boundaries and much reduced computational load.

    Let $u$ and $v$ be adjacent nodes of the current segmentation, and let $b_{u, v}$ be the boundary separating them.
    From each cue described above, we calculated the following features, which we concatenated into a single feature vector.

    \subsubsection{Pixel-level features}
        For $u$, $v$, and $b_{u, v}$, we created a histogram of 10 or 25 bins, and computed 3 or 9 approximate quantiles by linear interpolation of the histogram bins.
        We also included the number of pixels, the mean value and 3 central moments.  
        Additionally, we used the differences between the central moments of $u$ and $v$, and the Jensen-Shannon divergence between their histograms.

    \subsubsection{Mid-level features}
        For natural image segmentation, we added several mid-level features based on region orientation and convex hulls.
        For orientation features, the orientation of each region is estimated from the region's second moment matrix.
        We use the angle between the two regions, as well and the angles between each region and a line segment connecting their centroids, as features.
        For convex hull features, we calculated the volume of the convex hull of each region, as well as for their union, and used the ratios between these convex hulls volumes and the volumes of the regions themselves as a measure of the convexity of regions.

%%% RESULTS AND DISCUSSION %%%
% \input{results.tex}

\section{Results}

\subsection{Evaluation}
\label{sec:evaluation}
    Before we describe the main results of our paper, a discussion of evaluation methods is warranted, since even the question of the ``correct'' evaluation method is the subject of active research.

    The most commonly used method is boundary precision-recall \cite{martin_learning_2004, arbelaez10}.
    A test segmentation and a gold standard can be compared by finding a one-to-one match between the pixels constituting their segment boundaries.
    Then, matched pixels are defined as true positives (TP), unmatched pixels in the automated segmentation are false positives (FP), and unmatched pixels in the gold standard are false negatives (FN).
    A measure of closeness to the gold standard is then given by the precision and recall values, defined as $P = TP/(TP+FP)$ and $R = TP/(TP+FN)$.
    The precision and recall can be combined into a single score by the F-measure, $F = 2PR/(P + R)$.
    A perfect segmentation has $P = R = F = 1$.

    The use of boundary precision-recall has deficiencies as a segmentation metric, since small changes in boundary detection can result in large topological differences between segmentations.
    This is particularly problematic in neuronal EM images, where the goal of segmentation is to elucidate the connectivity of extremely long, thin segments that have tiny (and error-prone) branch points.
    For such images, the number of mislabeled boundary pixels is irrelevant compared to the precise location and topological impact of the errors \cite{Jain:2010,Turaga:2009}.
    In what follows, we shall therefore focus on region-based metrics, though we will show boundary PR results in the context of natural images to compare to previous work.

    The region evaluation measure of choice in the segmentation literature has been the Rand index (RI) \cite{rand_objective_1971}, which evaluates pairs of points in a segmentation.
    For each pair of pixels, the automatic and gold standard segmentations agree or disagree on whether the pixels are in the same segment.
    RI is defined as the proportion of point pairs for which the two segmentations agree.
    Small differences along the boundary have little effect on RI, whereas differences in topology have a large effect.

    However, RI has several disadvantages, such as being sensitive to rescaling and having a limited useful range \cite{vinh_information_2010}.
    An alternative segmentation distance is the variation of information (VI) metric \cite{meila_comparing_2003}, which is defined as a sum of the conditional entropies between two segmentations:

    \begin{equation}
    VI(S,U) = H(S | U) + H(U | S),
    \label{eq:vi}
    \end{equation}

    where $S$ is our candidate segmentation and $U$ is our ground truth.
    $H(S|U)$ can be intuitively understood as the answer to the question: ``given the ground truth (U) label of a random voxel, how much more information do we need to determine its label in the candidate segmentation (S)?''

    VI overcomes all of the disadvantages of the Rand index and has several other advantages, such as being a formal metric \cite{meila_comparing_2003}.
    Although VI has been used for evaluating natural image segmentations \cite{arbelaez10}, its use in EM has been limited.
    In what follows, we explore further the properties of VI as a measure of segmentation quality and conclude that it is superior to the Rand index for this task, especially in the context of neuronal images.

    Like the Rand index, VI is sensitive to topological changes but not to small variations in boundary changes, which is critical in EM segmentation.
    Unlike RI, however, errors in VI scale linearly in the size of the error whereas the RI scales quadratically.
    This makes VI more directly comparable between volumes.
    In addition, because RI is based on point pairs, and because the vast majority of pairs are in disjoint regions, RI has a limited useful range very near 1, and that range is different for each dataset.
    In contrast, VI ranges between 0 and $\log(K)$, where $K$ is the number of objects in the image.
    Furthermore, due to its basis in information theory, it is measured in bits, which makes it easily interpretable.
    For example, a VI value of 1 means that on average, each neuron is split in 2 equally-sized fragments in the automatic segmentation (or vice-versa).
    No such mapping exists between RI and a physical intuition.
    Finally, because VI is a metric, differences in VI correspond to our intuition about distances in Euclidean space, which allows easy comparison of VI distances between many candidate segmentations.

\begin{figure}[h]
    \begin{center}
    \includegraphics[width=3.27in]{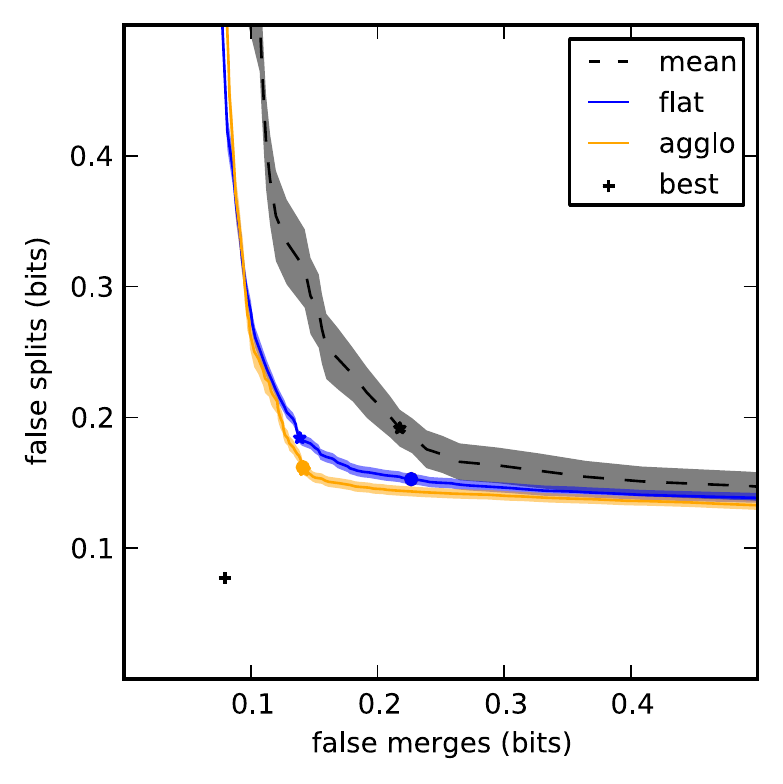}
    \end{center}
    \caption{{\bf Split VI plot for different learning or agglomeration methods.}
        Shaded areas correspond to mean $\pm$ standard error of the mean.
        ``Best'' segmentation is given by optimal agglomeration of superpixels by comparing to the gold standard segmentation.
        This point is not $(0,0)$ because the superpixel boundaries do not exactly correspond to those used to generate the gold standard.
        The standard deviation of this point ($n=8$) is smaller than the marker denoting it.
        Stars mark minimum VI (sum of false splits and false merges), circles mark VI at threshold 0.5.
    }
    \label{fig:fibsem-split-vi}
    \end{figure}

    VI is by its definition (Equation \ref{eq:vi}) broken down into an oversegmentation/false-split term $H(S | U)$ and an undersegmentation/false-merge term $H(U | S)$.
    To make this explicit, we introduce in this work the split-VI plot of $H(S | U)$ on the y-axis against $H(U | S)$ on the x-axis, which shows the tradeoff between oversegmentation and undersegmentation in a manner similar to boundary PR curves (see Figures \ref{fig:fibsem-split-vi} and \ref{fig:natural-compare}).
    Since VI is the sum of those two terms, isoclines in this plot are diagonal lines sloping down.
    A slope of $-1$ corresponds to equal weighting of under- and oversegmentation, while slopes of $-a$ correspond to a weighting of $a$ of undersegmentation relative to oversegmentation.
    Finding an optimal segmentation VI is thus as easy as finding a tangent for a given curve.
    The split-VI plot is particularly suited to agglomerative segmentation strategies: the merging of two segments can only result in an arc towards the bottom-right of the plot;
    false merges result in mostly rightward moves, while true merges result in mostly downward moves.

    In addition, each of the under- and oversegmentation terms can be further broken down into its constituent errors.
    The oversegmentation term of a VI distance is defined as $H(S|U) = -\sum_{u}{P(u) H(S|U=u) }$.
    From this definition, we introduce the VI breakdown plot, of $H(S|U=u)$ against $P(U=u)$ for every value of $u$, and vice-versa.
    In Supplementary Figure S1, we show how this breakdown can be used to gain insight into the errors found in automatic segmentations by identifying those segments that contribute most to the VI.

    In light of the utility of VI, our evaluation is based on VI, particularly for EM data.
    For natural images, we also present boundary precision-recall and other measures, to facilitate comparison to past work.
    In addition to boundary PR values, RI, and VI, we show values for the covering, a measure of overlap between segments \cite{arbelaez10}.
    For each of these measures, we show results for the optimal dataset scale (ODS), the optimal image scale (OIS), and for the covering measure we also show the result of the best value using any threshold of the segmentation (Best). 
    For boundary evaluation, we also report the average precision (AP), which is the area under the PR curve.

\subsection{Algorithms}
    We present in this paper the segmentation performance of several agglomerative algorithms, defined below.
    As a baseline we show results from agglomeration using only the mean boundary probability between segments (``mean'').

    For natural images, we also show the results when oriented boundary maps are used (``mean-orient''), which is the algorithm presented by Arbel\'{a}ez \textit{et al.} \cite{arbelaez10} and was shown in their work to outperform previous agglomerative methods.
    (Our results vary slightly from those of Arbel\'{a}ez, due to implementation differences.)

    Our proposed method, using an actively-trained classifier and agglomeration, is denoted as ``agglo''.
    For details, see Section \ref{methods:agglo-learning} and Figure \ref{fig:agglo-learning}.
    Briefly, using a volume for which the true segmentation is known, we start with an initial oversegmentation, followed by an agglomeration step in which every merge is checked against the true segmentation.
    True merges proceed and are labeled as such, while false merges do not proceed, but are labeled as false.
    This accumulates a training dataset until the agglomeration matches the true segmentation.
    At this point, a new agglomeration order is determined by training, and the procedure is repeated a few times to obtain a large training dataset, the statistics of which will match those encountered during a test agglomeration.
    
    A similar method, described by Jain \textit{et al.} \cite{jain2011lash} is denoted as ``lash'' in Supplementary Figures S2 and S3.
    In that work, merges proceed regardless of whether they are true or false according to the ground truth, and each merge is labeled by taking the sign of the change in Rand index resulting from the merge.
    We used our own implementation of LASH, using our own feature maps, to compare only the performance of the learning strategies.

    In order to show the effect of our agglomerative learning, we also compare using a classifier trained on only the initial graph before agglomeration (``flat'').

\subsection{Segmentation of FIBSEM data}
\label{sec:em-results}
    Our starting dataset was a $500 \times 500 \times 500$ voxel isotropic volume generated by focused ion beam milling of \emph{Drosophila melanogaster} larval neuropil, combined with scanning electron microscope imaging of the milled surface \cite{Knott:2008bn}.
    This results in a volume with 10nm resolution in the x, y and z axes, in which cell boundaries, mitochondria, and various other cellular components appear dark (Figure \ref{fig:em-images}).
    Relative to other EM modalities, such as serial block face scanning EM (SBFSEM) \cite{Denk:2004ja} or serial section transmission EM (ssTEM) \cite{Hayworth:2006wr,Harris:2006hh}, FIBSEM has a smaller field of view, but yields isotropic resolution and can be used to reconstruct important circuits.
    Recently published work has demonstrated a $28 \times 28 \times$ \SI{56}{\micro\metre} volume imaged at $7 \times 7 \times$ \SI{7}{\nano\metre} resolution \cite{Lichtman:2011kh}, and the latest volumes being imaged exceed $65 \times 65 \times$ \SI{65}{\micro\metre} with 8nm isotropic voxels (C. Shan Xu and Harald Hess, pers. commun.).
    These dimensions are sufficient to capture biologically interesting circuits in the {\it Drosophila} brain, such as multiple columns in the medulla (part of the visual system) \cite{Takemura:2008ee} or the entire antennal lobe (involved in olfaction) \cite{Laissue:1999tn}.

    To generate a gold standard segmentation, an initial segmentation based on pixel intensity alone was manually proofread using software specifically designed for this purpose (called Raveler) \cite{Chklovskii:2010df}.
    We then used the 8 probability maps described in Section \ref{sec:features} in a cross-validation scheme, training on one of the 8 volumes and testing on the remaining 7, for a total of 56 evaluations per training protocol (but only 8 for mean agglomeration, which requires no training).

    Compared with mean agglomeration or with a flat learning strategy, our active agglomerative learning algorithm improved segmentation performance modestly but significantly (Figure \ref{fig:fibsem-split-vi}).

\begin{figure*}[!ht]
    \begin{center}
    \includegraphics[width=6.83in]{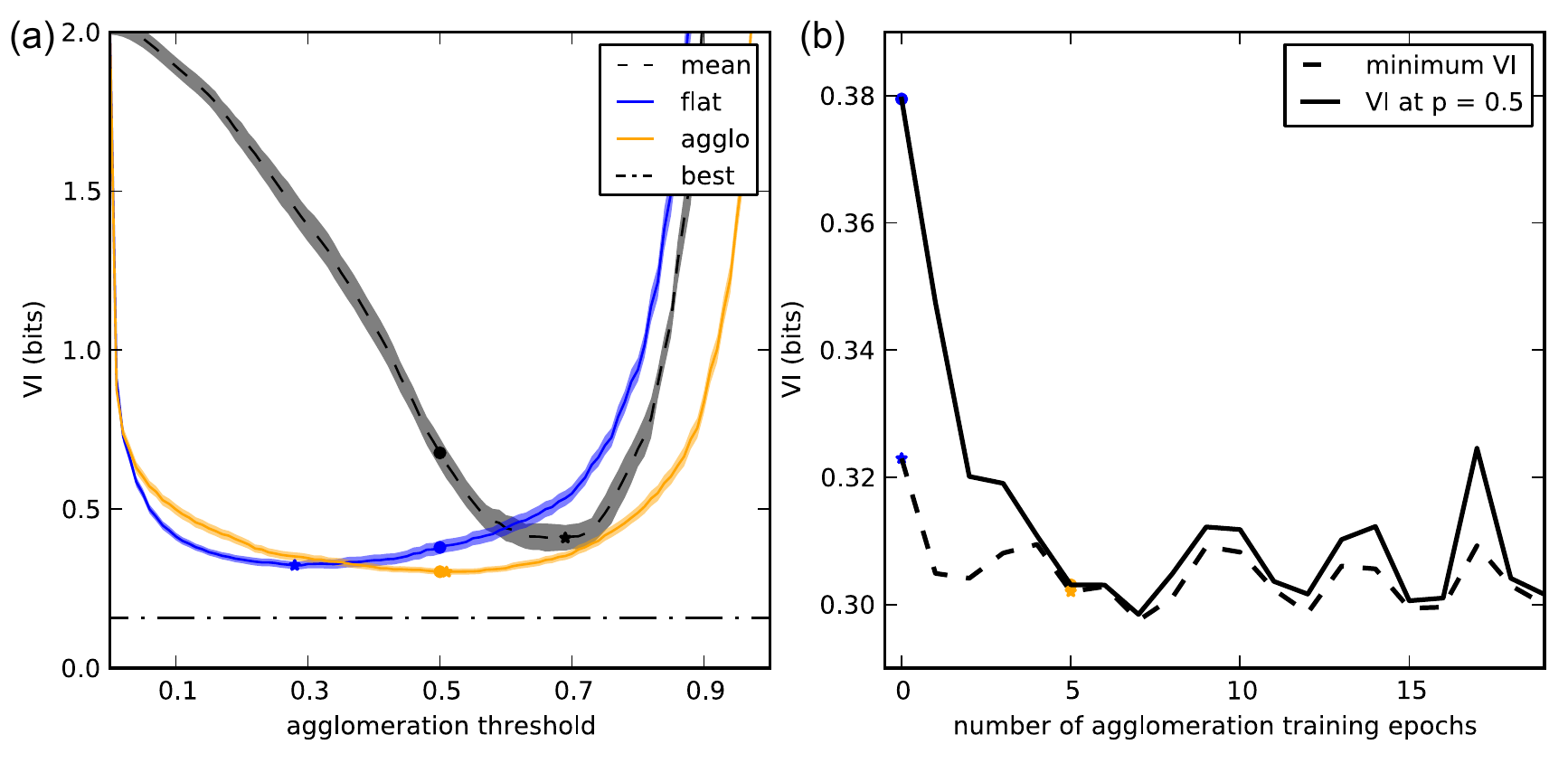}
    \end{center}
    \caption{{\bf Agglomerative learning improves merge probability estimates during agglomeration. \bf}
        (Flat learning is equivalent to 0 agglomerative training epochs.)
        (a) VI as a function of threshold for mean, flat learning, and agglomerative learning (5 epochs). Stars indicate minimum VI, circles indicate VI at $p=0.5$.
        (b) VI as a function of the number of training epochs.
        The improvement in minimum VI afforded by agglomerative learning is minor (though significant), but the improvement at $p=0.5$ is much greater, and the minimum VI and VI at $p=0.5$ are very close for 4 or more epochs.
    }
    \label{fig:fibsem-epochs}
    \end{figure*}

    In addition, the agglomerative training appears to dramatically improve the probability estimates from the classifier.
    If the probability estimates from a classifier are accurate, then, under reasonable assumptions, we expect the minimum VI to occur at or near $p=0.5$.
    However, this is not what occurs after learning on the flat graph: the minimum occurs much earlier, at $p=0.28$, after which the VI starts climbing.
    In contrast, after agglomerative learning, the minimum does indeed occur at $p=0.51$ (Figure \ref{fig:fibsem-epochs}a).

    This suggests that agglomerative learning improves the classifier probability estimates.
    Indeed, the minimum VI and the VI at $p=0.5$ converge after 4 agglomerative learning epochs and stay close for 19 epochs or more (Figure \ref{fig:fibsem-epochs}b).
    This accuracy can be critical for downstream applications, such as estimating proofreading effort \cite{Plaza:2012vo}.

\subsection{Segmentation of the SNEMI3D challenge data}
    Although we implemented our algorithm to work specifically on isotropic data, we attempted to segment the publicly available SNEMI3D challenge dataset (available at \url{http://brainiac2.mit.edu/SNEMI3D}), a $6 \times 6 \times$ \SI{30}{\nano\metre} resolution serial section scanning EM (ssSEM) volume.
    For this, we used the provided boundary probability maps of Ciresan {\it et al.} \cite{Ciresan:2012vi}.
    A fully 3D workflow, including 3D watershed supervoxels, predictably did not impress (adjusted Rand error 0.335, placed 3rd of 4 groups, 15th of 21 attempts).
    However, with just one modification (generating watershed superpixels in each plane separately), running GALA out of the box in 3D placed us in 1st place (as of this submission), with an adjusted Rand error of 0.125.
    (Note: our group name in the challenge is ``FlyEM''. To see individual submissions in addition to group standings, it is necessary to register and log in.)
    This demonstrates that the GALA framework is general enough to learn simultaneous 2D segmentation and 3D linkage, despite its focus on fully isotropic segmentation.
    We expect that the addition of linkage-specific features would further improve GALA's performance in this regime.

\subsection{Berkeley Segmentation Dataset} 
    We also show the results of our algorithm on the Berkeley Segmentation Dataset (BSDS500) \cite{arbelaez10}, a standard natural image segmentation dataset, and show a significant improvement over the state of the art in agglomerative methods.

    Our algorithm improves segmentation as measured by all the above evaluation metrics (Table \ref{tab:BSDS500}).
    At the optimal dataset scale (ODS), our algorithm reduced the remaining error between oriented mean agglomeration \cite{arbelaez10} and human-level segmentation by at least 20\% for all region metrics, including a reduction of 28\% for VI.
    The improvement obtained by agglomerative learning over flat learning is smaller than in EM data; we believe this is due to the smaller range of scales found between superpixels and segments in our natural images.
    Nevertheless, this slight improvement demonstrates the advantage of our learning method: by learning at all scales, the classifier achieves a better segmentation since it can dynamically adjust how features are interpreted based on the region size.

    \begin{table*}[!ht]
    \centering
    \caption{Evaluation on BSDS500. Higher is better for all measures except VI, for which lower is better. ODS uses the optimal scale for the entire dataset while OIS uses the optimal scale for each image.}
    \subtable[Region evaluation]{
    \begin{tabular}{l|lll|ll|ll|}
    \toprule
     \multicolumn{1}{c}{\bf Algorithm} & \multicolumn{3}{c}{\bf Covering} & \multicolumn{2}{c}{\bf RI} & \multicolumn{2}{c}{\bf VI}\\
    & \bf ODS & \bf OIS & \bf Best & \bf ODS & \bf OIS & \bf ODS & \bf OIS \\
    \midrule
    human & 0.72 & 0.72 & | & 0.88 & 0.88 & 1.17 & 1.17\\
    \midrule
    agglo & \bf0.612 & \bf0.669 & \bf0.767 & \bf0.836 & \bf0.862 & \bf1.56 & \bf1.36  \\
    flat & 0.608 & 0.658 & 0.753 & 0.830 & 0.859 & 1.63 & 1.42\\
    oriented mean \cite{arbelaez10} &0.584 & 0.643 & 0.741 & 0.824 & 0.854 & 1.71 & 1.49\\
    mean & 0.540 & 0.597 & 0.694 & 0.791 & 0.834 & 1.80 & 1.63 \\
    \end{tabular}
    }
    \subtable[Boundary evaluation]{
    \begin{tabular}{|lll}
    \toprule
    \multicolumn{3}{c} {\bf F-measure}\\
     \multicolumn{1}{c} {\bf ODS}&\bf  OIS &\bf  AP\\
     \midrule
    0.80 &0.80 &|\\
    \midrule
     \bf0.728 & \bf0.760 & \bf0.777  \\
     0.726 & \bf0.760 & 0.776\\
     0.725 & 0.759 & 0.758\\
    0.643 & 0.666 & 0.689 \\
    \end{tabular}
    }
    \label{tab:BSDS500}
    \end{table*}

\begin{figure*}[!ht]
    \begin{center}
    \includegraphics[width=6.83in]{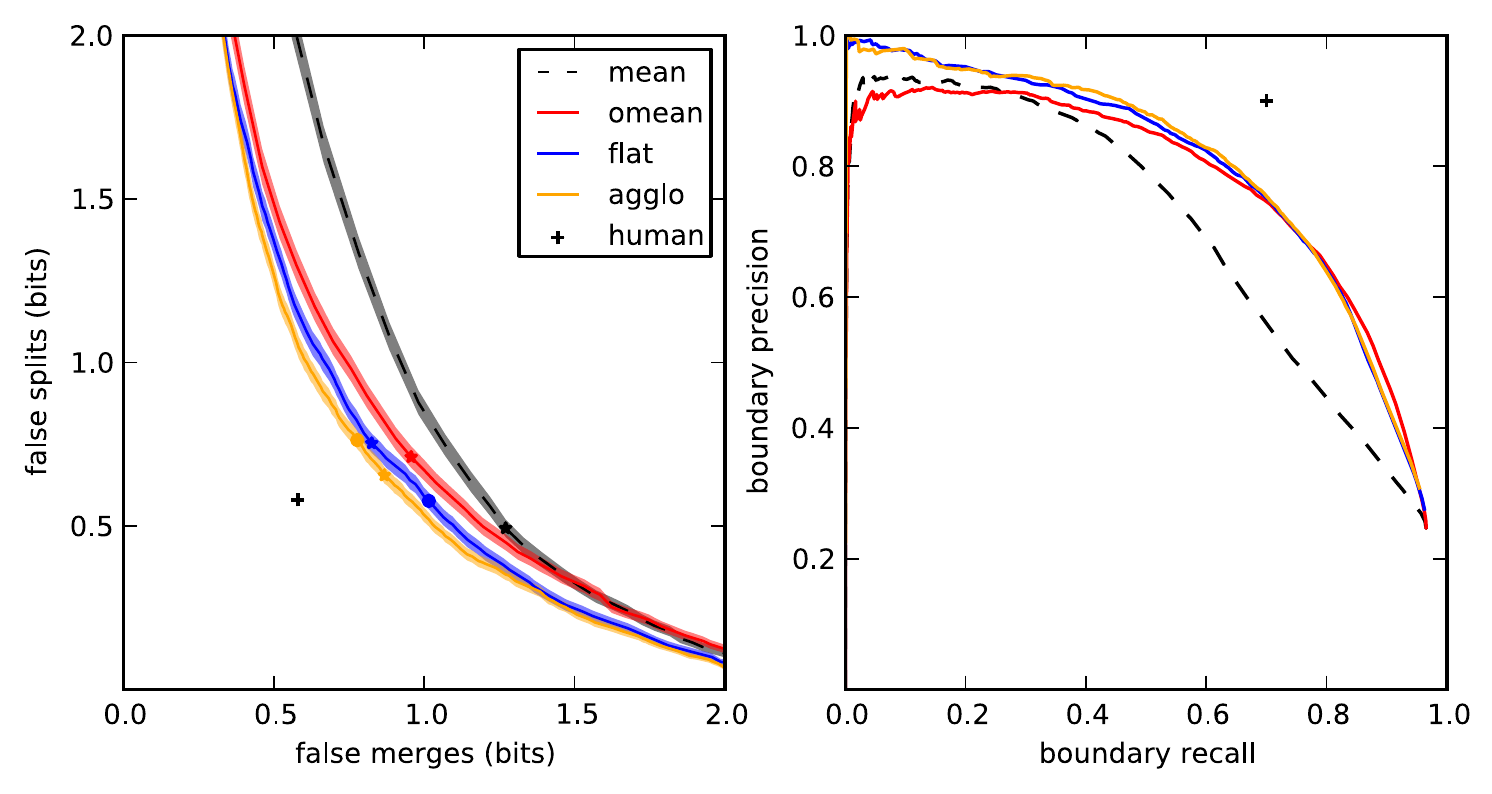}
    \end{center}
       \caption{\bf Evaluation of segmentation algorithms on BSDS500.}
    \label{fig:natural-compare}
    \end{figure*}

    Figure \ref{fig:natural-compare}a shows the split VI plot while Figure \ref{fig:natural-compare}b shows the boundary precision-recall curves. 
    The results are similar in both cases, with agglomerative learning outperforming all other algorithms.

\begin{figure}[!ht]
    \begin{center}
    \includegraphics[width=3.27in]{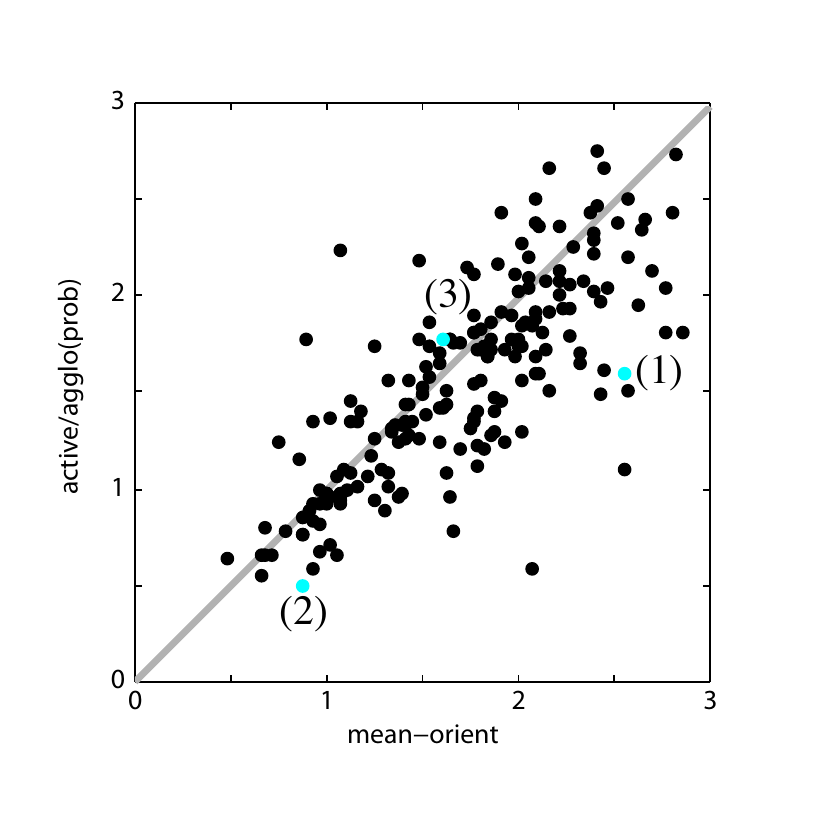}
    \end{center}
       \caption{{\bf Comparison of oriented mean and actively learned agglomeration.} as measured by VI at the optimal dataset scale (ODS).
       Each point represents one image.
       Numbered and colored points correspond to the example images in Figure \ref{fig:natural-examples}.}
    \label{fig:scatter-natural}
    \end{figure}

    In Figure \ref{fig:scatter-natural}, we show the performance of our algorithm on each test image compared to the algorithm in \cite{arbelaez10}.
    The majority of test images show a better (i.e. lower) VI score.

\begin{figure*}
    \begin{center}
    \includegraphics[width=6.83in]{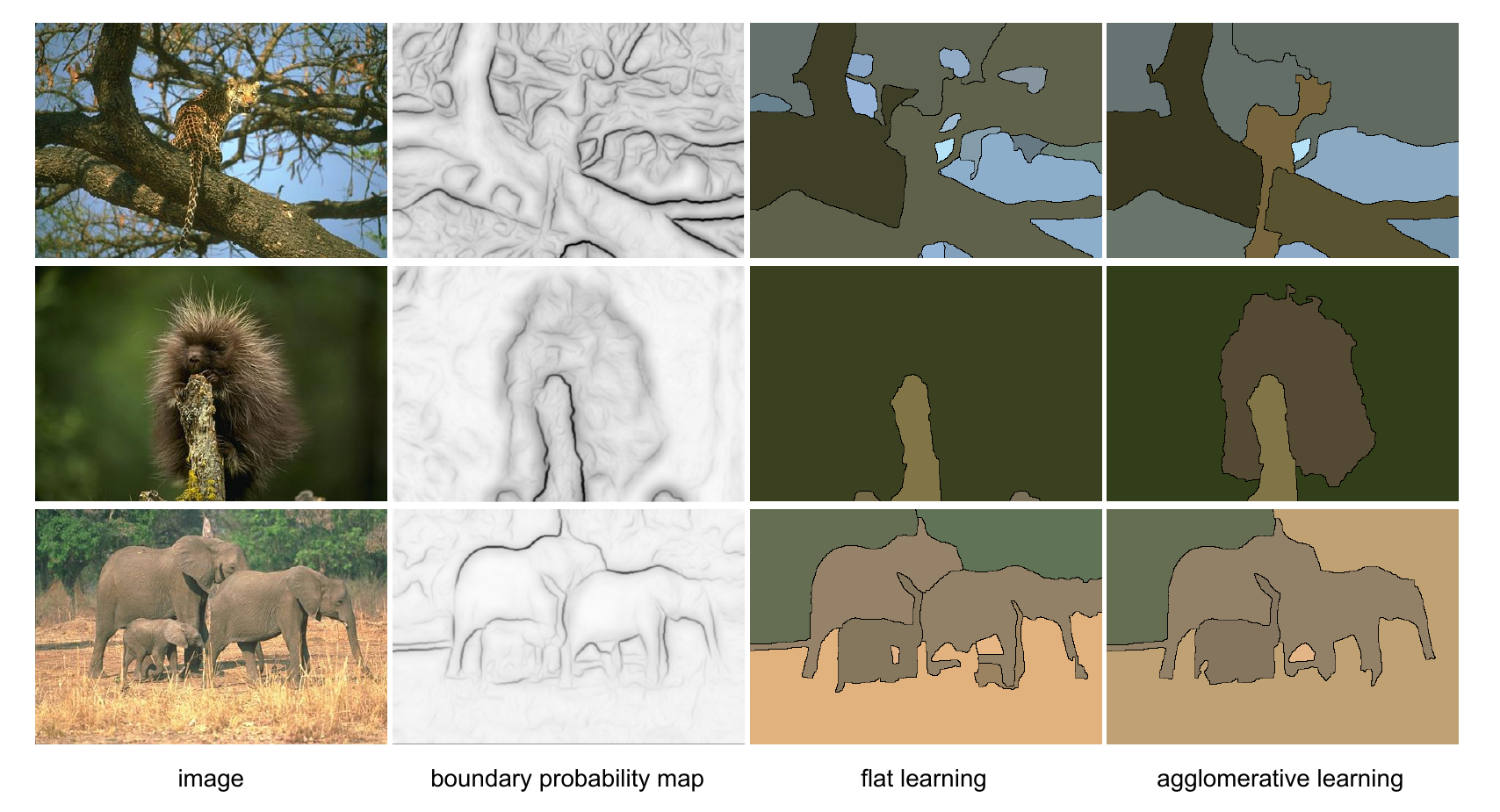}
    \end{center}
    \caption{{\bf Example segmentations on natural images}.
        {\it Top row}: Despite having a very noisy boundary map, using additional cues allows us to segment the objects successfully.
        {\it Middle row}: Although there are many weak edges, region-based texture information helps give a correct segmentation.
        {\it Bottom row}: A failure case, where the similar texture of elephants causes them to be merged even though a faint boundary exists between them.
        For all rows, the VI ODS threshold was used.
        The rows correspond top to bottom to the points identified in Figure \ref{fig:scatter-natural}.}
    \label{fig:natural-examples}
    \end{figure*}

    Several example segmentations are shown in Figure \ref{fig:natural-examples}.
    By learning to combine multiple cues that have support on larger, well-defined regions, we are able to successfully segment difficult images even when the boundary maps are far from ideal.

\section{Discussion and conclusions}
\label{sec:discussion}
    We have presented a method for learning agglomerative segmentation.
    By performing agglomeration while comparing with a ground truth, we learn to merge segments at all scales of agglomeration.
    And, by guiding the agglomeration with the previous best policy, we guarantee that the examples we learn match those that will be encountered during a test agglomeration.
    Indeed, the difference in behavior between agglomerative learning and flat learning is immediately apparent and striking when watching the agglomerations occur side by side (see Supplementary Video S4).

    LASH \cite{jain2011lash} is a similar approach to ours that has nonetheless important conceptual differences.
    We use our gold standard segmentation to guide agglomeration during learning --- preventing false merges --- while they follow their current policy to completion, and use the sign of the change in Rand index as the learning label.
    A case can be made for either approach: in our case, we can train merges and non-merges from correct segments of arbitrary size, while LASH might diverge from the correct segmentation early on and then essentially train on noisy segments.
    We have anecdotally observed this advantage in play when we successfully used training data from a $250^3$ voxel volume to segment a $500^3$ voxel test volume.
    On the other hand, our own classifier might not get suitable training data for the times it diverges from a correct segmentation.
    Mixed training datasets from both strategies could turn out to be the best approach, and we will explore this possibility in future work.
    
    Another difference is that Jain \textit{et al.} only keep the training data from the last training epoch, while we concatenate the data from all epochs.
    In our experiments, we saw a significant improvement, relative to LASH, in segmentation accuracy in natural image data (Supplementary Figure S2).
    In EM data, the improvement was still present but only at higher undersegmentation values (over-merging), with LASH displaying a smaller advantage earlier in the agglomeration (Supplementary Figure S3).
    
    Recent work also attempts to use machine learning to classify on a merge hierarchy starting from watershed superpixels \cite{Liu:2012ba}.
    Liu {\it et al.}'s method cleverly chooses the right watershed threshold locally by learning directly on the merge tree nodes.
    However, the algorithm uses a single hierarchy of watershed superpixels obtained with different merge thresholds.
    This means that errors in the original hierarchy cannot be corrected by the machine learning approach, and watershed thresholding has been previously shown to give poor segmentation results \cite{jain2011lash}.
    Our method, in contrast, updates the merge hierarchy after each training epoch, potentially rectifying any prior errors.
    Liu {\it et al.}'s novel use of merge potentials to dynamically find the optimal threshold in each branch of the hierarchy, however, could be useful in the case of GALA.
    
    Bjoern Andres, Fred Hamprecht and colleagues have devoted much effort to the use of graphical models to perform a one-shot agglomeration of supervoxels \cite{Andres:2011is,Andres:2012vp,Andres:2012ba,Andres:2008p7340}.
    Although they only learn region merge probabilities at the base level of supervoxels, their use of conditional random fields (CRFs) to find the most consistent merge configuration is an advantage that our greedy, hierarchical approach lacks.
    On the other hand, their approach has two distinct disadvantages, in scalability and proofreadability.

    First, the theoretical scalability of a global optimization is limited, which could become a problem as volumes exceed the teravoxel range.
    In contrast, GALA and other hierarchical methods could theoretically be implemented in a Pregel-like massively parallel graph framework \cite{Malewicz:2010to}, allowing the segmentation of extremely large volumes in time proportional to the number of supervoxels.
    
    Second, despite the significant progress of the last decade, the accuracy of all currently available segmentation methods is orders of magnitude too small for their output to be used directly without human proofreading \cite{Chklovskii:2010df,Jurrus:2013ef}.
    GALA operates locally, which makes proofreading possible because manually adding a cut or merge only affects a few nearby predictions.
    Furthermore, proofreading can occur on any of the scales represented by the hierarchy.
    In contrast, because of the global optimization associated with the CRF approach, adding human-determined constraints to the supervoxel graph affects merge probabilities everywhere, resulting in expensive re-computation and the possibility that already-proofread areas need to be revisited.

    A lot of the effort in connectomics focuses on the segmentation of anisotropic serial-section EM volumes \cite{VazquezReina:2011uz,Laptev:2012cv,Funke:2012uv}.
    Much like Liu {\it et al.}, Vazquez-Reina {\it et al.} use watershed segmentations of boundary probability maps at multiple thresholds on each different plane of the serial-section stack.
    They then use a CRF to link segments from consecutive sections at potentially different watershed thresholds.
    Funke {\it et al.}, in contrast, use a superpixel-less approach to obtain simultaneous segmentation within planes and linkage between planes \cite{Funke:2012uv}.
    Their within-plane segmentation optimizes a segmentation energy term with smoothness constraints, which eliminates many of the weaknesses of watersheds.
    Although the separation of segmentation and linkage between sections is not necessary in isotropic datasets, these approaches could inspire extensions of GALA specifically aimed at anisotropic segmentation.

    The feature space for agglomeration is also worthy of additional exploration.
    For EM data, we included pixel probabilities of boundary, cytoplasm, mitochondria, and glia.
    Classifier predictions for synapses and vesicles might give further improvements \cite{Kreshuk:2011el}.
    Additionally, we found that most errors in our EM data are ``pinch'' errors, in which a neuronal process is split at a very thin channel.
    In these cases, features based on sums over voxels tend to be weakly predictive, because the number of voxels between the two segments is small.
    We are therefore actively exploring features based on segment shape and geometry, which have indeed been very useful in the work of Andres {\it et al.} discussed above \cite{Andres:2011is,Andres:2012vp,Andres:2012ba,Andres:2008p7340}.
    Furthermore, we note that community-standard implementation of features will aid in the comparison of different learning and agglomeration algorithms, which are at present difficult to evaluate because they are conflated with the feature computation.
    A direct comparison of the segmentation performance of CRFs and agglomerative methods, disentangled from feature maps, would serve to advance the field.

    A weakness of our method is its requirement for a full gold standard segmentation for training.
    This data might not be easily obtained, and indeed this has been a bottleneck in moving the method ``from benchside to bedside'', so to speak.
    We are therefore in the process of modifying the method to a semi-supervised approach that would require far less training data to achieve similar performance.
    
    Finally, the field of neuronal reconstruction will depend on segmentation algorithms that not only segment well, but point to the probable location of errors.
    Although it requires further improvements in speed, scalability, and usability, our method is a first step in that direction.

\section*{Data and code availability}
The source code for the Gala Python library can be found at:

\url{https://github.com/janelia-flyem/gala}.

The EM dataset presented here in this work can be found at:

\url{https://s3.amazonaws.com/janelia-free-data/Janelia-Drome-larva-FIBSEM-segmentation-data.zip}.

%%% ACKNOWLEDGEMENTS %%%
\section*{Acknowledgements}
We thank Bill Katz for critical reading of the manuscript, C. Shan Xu and Harald Hess for the generation of the image data, Mat Saunders for generation of the ground truth data, Shaul Druckmann for help editing figures, and Viren Jain, Louis Scheffer, Steve Plaza, Phil Winston, Don Olbris and Nathan Clack for useful discussions.

\bibliography{references}

\end{document}